# IoT Botnet Detection: Application of Vision Transformer to Classification of Network Flow Traffic

Hassan Wasswa
*School of Engineering and Information Technology*
*University of New South Wales*
Canberra, Australia
h.wasswa@adfa.edu.au

Timothy Lynar
*School of Engineering and Information Technology*
*University of New South Wales*
Canberra, Australia
t.lynar@adfa.edu.au

Aziida Nanyonga
*School of Engineering and Information Technology*
*University of New South Wales*
Canberra, Australia
a.nanyonga@adfa.edu.au

Hussein Abbass
*School of Engineering and Information Technology*
*University of New South Wales*
Canberra, Australia
h.abbass@unsw.edu.au

*Abstract*—Despite the demonstrated effectiveness of transformer models in NLP, and image and video classification, the available tools for extracting features from captured IoT network flow packets fail to capture sequential patterns in addition to the absence of spatial patterns consequently limiting transformer model application. This work introduces a novel preprocessing method to adapt transformer models, the vision transformer (ViT) in particular, for IoT botnet attack detection using network flow packets. The approach involves feature extraction from .pcap files and transforming each instance into a 1-channel 2D image shape, enabling ViT-based classification. Also, the ViT model was enhanced to allow use any classifier besides Multilayer Perceptron (MLP) that was deployed in the initial ViT paper. Models including the conventional feed forward Deep Neural Network (DNN), LSTM and Bidirectional-LSTM (BLSTM) demonstrated competitive performance in terms of precision, recall, and F1-score for multiclass-based attack detection when evaluated on two IoT attack datasets.

## I. INTRODUCTION

The rapid advancement of Internet of Things (IoT) technology is driving an unparalleled transformation in cyber-physical systems and with unprecedented benefits to users. The number of IoT devices is exponentially rising and envisaged to reach 29.24 billion by 2030 according to Statista 2021 report. This substantial quantity has significantly influenced various domains of applications, including but not limited to healthcare, transportation, smart homes, urban infrastructure, and energy management.

However, this rapid technological advancement has not been matched by adequate focus on IoT security. As a result, these devices are susceptible to attacks which has led to the rising popularity of IoT as a potent instrument for cybercriminals. The most popular avenue for executing IoT-based attacks is by exploiting vulnerable IoT devices and add them to IoT botnets. The IoT botnets are then used to launch various sorts of distributed attacks like massive DDoS, fraud clicks, phishing, distributed malware, distributed spam attacks, etc. Notably, the Mirai botnet, composed of roughly 100,000 compromised IoT devices (including smart routers, smart DVRs, and smart IP cameras), conducted massive DDoS attacks that crippled operations of various companies including Dyn, Amazon, Twitter, and Reddit in 2016 [1]. And ever since the Mirai source code was made open to public access in 2017, a large number of its variants has emerged in the recent years. With this continuous emergence of more advanced malware versions, designed to establish large-scale botnets, advanced and effective attack detection techniques have become pivotal in preventing botnet attacks.

To stay abreast of the challenge and owing to the recent success of AI-based models in solving regression, binary and multi-class classification problems [2], [3], a concerted effort has been invested in the design of efficient and reliable AI-based solutions for detection of IoT driven attacks. Various studies have proposed methods for detection of IoT botnet attacks ranging from the use of conventional classification algorithms like Naive Bayes, Decision Trees, Random forest, AdaBoost, ExtraTree, and k-Nearest neighbors [4], [5], to deep learning based techniques with many layers of com-plex mathematical operations ranging from conventional feed forward DNN algorithms to advanced methods like Convolutional Neural Networks (CNNs) [6], [7], Recurrent Neural networks (RNNs) [8], [9]. In addition to deployment of advanced classification algorithms, advanced deep learning-based data generation and preprocessing techniques including Auto-Encoders (AE), variational Auto-Encoder (VAE), and Generative Adversarial Networks (GAN) based techniques, have been proposed and deployed for enhancing the performance of detection models [10]–[12].

However, despite the demonstrated effectiveness of transformer models [13] in NLP, and the vision transformer

(ViT) [14] for image and video classification, the application of transformers-based architectures to network flow packets for attack detection has been limited due to lack of both sequential and spatial patterns exhibited in text and image datasets, respectively. To this end, this work proposes a data preprocessing approach that transforms network flow-based packet instances (.pcap) to a format that can be processed by the ViT for IoT botnet attack detection. The approach involves extraction of features from .pcap files, transforming each extracted flow instance features to a 1-channel 2D image shape which can then be processed by the ViT-based model for classification. Also, unlike the conventional ViT model which was based on MLP only for classification, our approach allows any classifier to be stacked on top of the ViT encoder model for both regression and classification purposes. Evaluated on two IoT datasets, the proposed model recorded competitive performance in terms of precision, recall and F1-score in a multi-class detection problem. The contribution of this work is two-fold:

1) Dynamic detection of NetFlow packets: By using the proposed model, attacks can be detected as data packets enter the IoT network. These packets can be captured in real-time, converted into 2D images, fed into the transformer model for feature extraction, and then categorized as either harmless or belonging to various IoT attack categories.
2) In contrast to previous research that restricted ViT's classifier component to the MLP model for classification, this study enhances the ViT model to permit use of any classifiers ranging from basic artificial neural networks (ANN) to more complex architectures like feed forward DNN, and RNN-based models including LSTM, and BLSTM models.

## II. RELATED WORK

An enormous amount of prior research has been conducted with a purpose of enhancing the IoT botnet attack detection efficiency. However, the self-attention based transformer architecture model has not seen extensive application as far as the IoT botnet domain is concerned despite its outstanding performance in areas of machine translation [15], [16], text classification [17], [18] and computer vision. This can be attributed to the fact that the transformer model was initially designed for NLP tasks where the training features exhibit sequential patterns while the ViT was initially designed for images which exhibit spatial structures. Unlike text and image datasets, NetFlow features extracted from network captures, by convention, neither exhibit sequential patterns nor spatial structures greatly hindering the use of transformer models for classification of such dataset instances.

The study in [19] proposed an approach for detection of cross-site scripting (XSS) attacks and compared the performance of a Transformer architecture-based model against two RNN models including LSTM and GRU. Each url was segmented into its constituent text components to create sequences of words. In order to train the transformer, natural language processing methods such as tokenization, word2vec with skip-gram model, TF-IDF and CBOW were deployed to create and represent word feature vectors. The authors evaluated their approach on datasets used in studies prior to their work and urls from the Xssed.com website and reported that the Transformer model had the potential to enhance XSS attack detection accuracy.

In study [20], proposed an approach for classification of IoT botnet malware using transformer model. Their experiment involved disassembling the malware opcodes of two known malware classes including Mirai and Gafgyt (commonly known as BASHLITE), one unknown IoT malware and benign samples. The authors implemented and utilized a script to retrieve the instruction sequences of the opcode functions and deployed CBOW for word vector representations. The proposed Transformer model was the used to extract features and a feed forward neural network for classification. The proposed model recorded competitive detection performance of up 99.12% average recall rate and 94.67% average malware classification accuracy.

The ViT model was deployed in study [21] for image spoofing attack detection in face recognition systems. The transformer was deployed to extract fine face features from public face image spoofing datasets including spoofing in the wild (SiW) [22] and Oulu-NPU [23] datasets. To improve the detection of unseen images, three forms of patch-wise data augmentation were proposed and implemented including intra-class patch mixing, live patch mask and patch cutout data augmentation. The approach outperformed existing state-of-the-art methods in terms of APCER, BPCER and ACER.

In study [24] a DGA domain detection scheme was pro-posed for detection of fake domain names generated and used by botnet devices to communicate with the command-and-control (C&C) server in a botnet attack. The study utilised a pretrained transformer (CANINE transformer model) architecture-based model to discriminate between benign do-main names and DGA-generated domain names. To enable the domain names to be processed by the CANINE transformer model, each domain name was tokenized into individual character sequences, and the three required input components (input ids, attention mask, and token type id) were created and used to fine-tune the transformer model. Evaluated on 1 million DGA-domains and 1 million benign domain names obtained from public sources, the proposed scheme outperformed existing state-of-the-art methods with accuracy, precision, recall and F1-score all reaching up to 99%.

However, all the above studies focus on unstructured dataset either in form of text sequences or images in a static form. This work differs from prior studies in two fundamental aspects:

1) Dynamic detection of NetFlow packets: The work in this study can be directly deployed to detect attacks as packets flow into the IoT network. The packets can be captured on the fly, transformed into 1-channel 2D images, passed into the transformer model for feature extraction and then classified as either benign or one of the IoT attack classes by a pre-trained classifier stacked on the output layer of the ViT encoder.

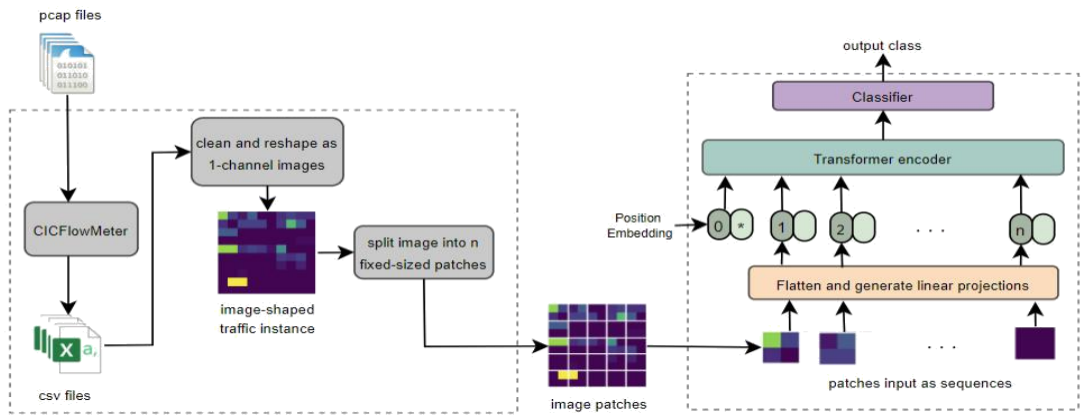

Fig. 1: Proposed vision transformer-based Attack detection model from flow packets

2) Stacking various deep learning models on top of the transformer encoder: All prior studies have limited the classifier component to only the multilayer perceptron (MLP) model for instance classification. This work implements an approach which allows any kind of classifier ranging from a simple ANN to deep feed-forward NN, to RNN based models including LSTM, BLSTM, and GRU. We compare the detection performance of different deep learning algorithms when stacked on top of the ViT encoder component.

## III. PROPOSED APPROACH

The approach presented in this study intends to adapt the ViT model for NetFlow traffic classification to facilitate dynamic detection of IoT Botnet attacks. It transforms the captured flow packets (.pcaps) into image representations before feeding them into the ViT encoder model for feature extraction and dimension reduction. Subsequently, the ViT encoder output is fed into a classifier for traffic classification. That is, the input layer of the classification algorithm is directly connected to the output layer of the ViT encoder model, as illustrated in Figure 1. Also, unlike the original ViT architecture that was fine tuned for the standard MLP classifier, our approach enhances the model to work with any classification algorithm.

### A. Transformer model

The Transformer architecture, which was proposed in study [13], brought about a revolutionary shift in the realm of NLP tasks by achieving exceptional outcomes. Contrast to conventional RNN models, the Transformer employs a multi-head self-attention mechanism to enable parallel processing. Consequently, the slow sequential training limitations associated with conventional RNN models are mitigated. The self-attention mechanism not only enhances computational speed but also, adeptly captures intricate dependencies among various components. The core principle of the attention mechanism involves the mapping of queries and key-value pairs to generate an output. This output is derived through a weighted summation of values, with the magnitudes of these weights determined by the compatibility between queries and their corresponding keys.

Figure 2 shows an extract of the transformer architecture as presented in study [13]

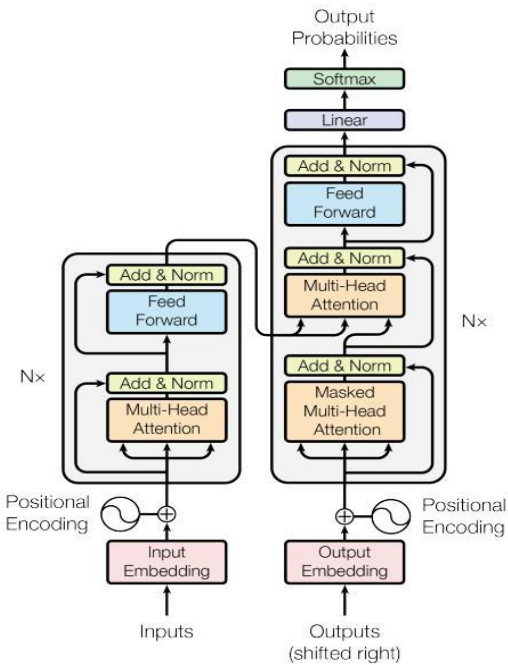

Fig. 2: Transformer model architecture as presented in Figure 1 of study [13]

### B. Vision Transformer

The Vision Transformer (ViT) was proposed in study [14] as an extension of the self-attention mechanism introduced in the conventional transformer model which was originally designed for NLP to enable image processing in computer vision tasks. Unlike traditional CNNs, ViT processes images by dividing them into fixed-size patches and using self-attention mechanisms to capture relationships among these patches. This enables ViT to efficiently understand global context and long-range dependencies in images. The fixed-sized patches are flattened into vectors and linearly embedded to create initial representations. To retain spatial information, ViT introduces positional embedding for each flattened patch vector enabling the model to understand the relative positions of patches within the image.

C. Dataset

The proposed approach was evaluated on two publicly available IoT botnet attack datasets: CIC-IoT 2022 [25] and Bot-IoT [26]. The CIC-IoT dataset, created by the Canadian Institute of Cybersecurity (CIC), contains benign and attack traffic captures in .pcap format. Employing the revised CI-CFlowMeter [1] tool, 87 traffic flow features were extracted, resulting in a dataset comprising over 3.2 million NetFlow records. The dataset's five multi-class labels (Normal, HTTP flood, TCP flood, UDP flood, and Brute force) were derived from directory names, and their instance distribution is presented in Table I.

On the other hand, the BoT-IoT dataset is publicly accessible for academic research purposes as .csv, .pcap or argus file. This work utilized the 5%-of-the-entire-dataset copy that was published as a .csv file. Considering that the "Theft" class of published 5% BoT-IoT dataset constituted only 79 data points, as shown in Table II, this class was dropped and only 4 classes including "DDoS", "DoS", "Reconnaissance", and "Normal" were considered for the multi-class classification problem.

TABLE I: CIC-IoT Dataset distribution

| Attack Category | No. of Instances | Percentage |
|---|---|---|
| Normal | 2,616,853 | 80.870 |
| HTTP flood | 554,316 | 17.130 |
| TCP flood | 45,884 | 1.418 |
| Brute force | 12,257 | 0.379 |
| UDP flood | 6,561 | 0.203 |

TABLE II: Bot-IoT Dataset distribution

| Attack Category | No. of Instances | Percentage |
|---|---|---|
| DDoS | 1,926,624 | 52.518 |
| DoS | 1,650,260 | 44.984 |
| Reconnaissance | 91,082 | 2.483 |
| Normal | 477 | 0.013 |
| Theft | 79 | 0.002 |

D. Data pre-processing

The data preprocessing involved removing missing values and dropping fields that were deemed irrelevant to model training. In both datasets, fields including IP addresses (both source and destination), packet number and sequence number/id were dropped. This was followed by reshaping each training instance to take the shape of a 1-channel 2D image. That is, each instance, $x \in R^d$ is expressed as $x' \in R^{r \times k \times 1}$ where $r \times k = d$. Each $x'$ is then split into equal sized image patches before it is fed into the ViT model. For cases where d is a prime number, zero-padding was introduced increasing the dimension d to d + 1 before the reshaping phase.

[1]https://github.com/GintsEngelen/CICFlowMeter

In this experiment, the CIC-IoT dataset comprised 84 independent variables after cleaning and each instance was reshaped as a $6 \times 14 \times 1$. Each reshaped instance was then split into 21 equal-sized $2 \times 2 \times 1$ image patches. On the other hand, the BoT-IoT dataset, after cleaning and dropping features including "pkSeqID", "flgs", "proto", "saddr", "daddr" and "state", constituted 37 independent variables. Since 37 is a prime number, each instance was padded with an extra 0-value making the dimension size 38. It should be noted that features including "flgs", "proto" and "state", were dropped since each had a corresponding numerically encoded field in the same dataset. Each of the instances was reshaped as a $2 \times 19 \times 1$ image and 19 equal-sized $2 \times 1 \times 1$ images patches were created from each of the image-shaped instances.

E. Auto-encoder model

For the purpose of performance comparison, a classical auto-encoder that projects a high dimensional dataset to a latent space of dimension 8 was trained on the Bot-IoT dataset. For a fair comparison the number of epochs, batch-size, learning rate and all other parameter settings were kept the same as those used to train the proposed ViT-based transformer model.

F. Learning models

Three different deep learning models were deployed for this study including a feed-forward Deep Neural Network (DNN) constituting two hidden layers, a LSTM model, and a BLSTM model. The LSTM, and BLSTM models were implemented by replacing the first hidden layer of the DNN model with the respective model layer. Neuron activation utilized the ReLU function, except for the output layer which used softmax. During model training, the Adam optimizer was utilized, and the categorical cross-entropy loss served as the loss function. Additionally, a 20%-random neuron dropout was deployed to curb overfitting. For all models, 10 epochs were utilized during training.

IV. RESULTS AND DISCUSSION

The experimental findings of this study are presented and discussed. As already indicated in section III-C, two datasets were utilised in this study. Table III shows the classification performance of the three classification algorithms deployed in this study, including DNN, LSTM, and BLSTM, in terms of Precision, Recall and F1-score. It can be seen from Table III that the approach produced competitive results in terms of recognition of all five attack classes in the CIC-IoT 2022 dataset.

To better understand how the proposed approach compares against the traditional classical AE-based model, the performance of the three classifiers was evaluated against the highly imbalanced Bot-IoT dataset (see distribution in section III-C).

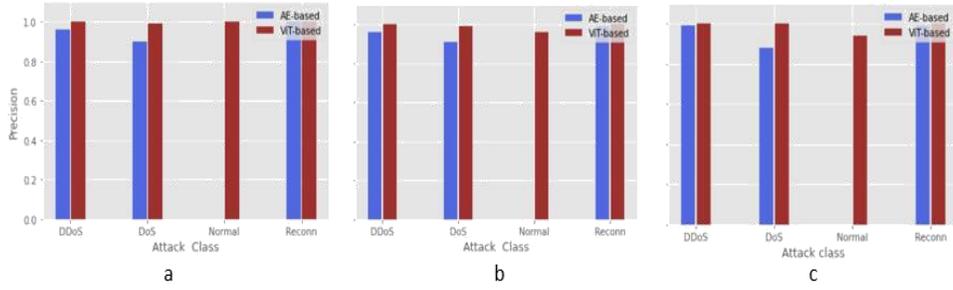

Fig. 3: AE-based versus proposed transformer-based model in terms of precision for: (a) DNN, (b) LSTM and (c) BLSTM

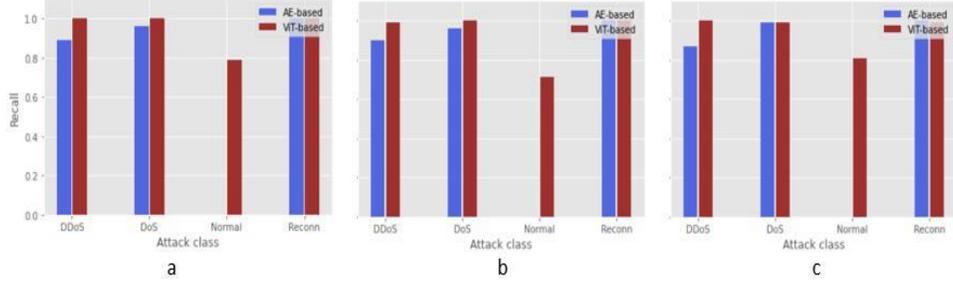

Fig. 4: AE-based versus proposed transformer-based model in terms of Recall for: (a) DNN, (b) LSTM and (c) BLSTM

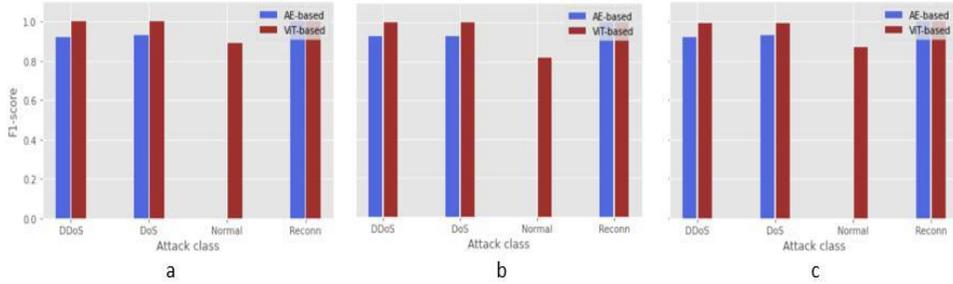

Fig. 5: AE-based versus proposed transformer-based model in terms of F1-Score for: (a) DNN, (b) LSTM and (c) BLSTM

TABLE III: Performance of the proposed approach on the CIC-IoT 2022 dataset in terms of Precision, Recall and F1-Score

| Model | Class | Precision | Recall | F1-Score |
|---|---|---|---|---|
| DNN | brute_force | 0.99 | 0.95 | 0.97 |
|  | http_flood | 0.90 | 1.00 | 0.95 |
|  | normal | 1.00 | 1.00 | 1.00 |
|  | tcp_flood | 0.98 | 0.52 | 0.68 |
|  | udp_flood | 1.00 | 0.94 | 0.97 |
| LSTM | brute_force | 0.97 | 0.67 | 0.79 |
|  | http_flood | 0.95 | 1.00 | 0.97 |
|  | normal | 1.00 | 1.00 | 1.00 |
|  | tcp_flood | 0.98 | 0.51 | 0.67 |
|  | udp_flood | 1.00 | 0.93 | 0.96 |
| BLSTM | brute_force | 0.98 | 0.91 | 0.95 |
|  | http_flood | 0.96 | 1.00 | 0.98 |
|  | normal | 1.00 | 1.00 | 1.00 |
|  | tcp_flood | 0.99 | 0.50 | 0.67 |
|  | udp_flood | 0.98 | 0.98 | 0.98 |

TABLE IV: Performance comparison between classical AE-based approach and proposed Transformer-based approach

| Model | Class | AE-based Approach | | | Proposed Approach | | |
|---|---|---|---|---|---|---|---|
|  |  | Prc | Recall | F1 | Prc | Recall | F1 |
| DNN | DDoS | 0.96 | 0.89 | 0.92 | 1.00 | 1.00 | 1.00 |
|  | DoS | 0.90 | 0.96 | 0.93 | 0.99 | 1.00 | 1.00 |
|  | Normal | 0.00 | 0.00 | 0.00 | 1.00 | 0.79 | 0.89 |
|  | Reconn | 1.00 | 1.00 | 1.00 | 1.00 | 1.00 | 1.00 |
| LSTM | DDoS | 0.96 | 0.90 | 0.93 | 1.00 | 0.99 | 1.00 |
|  | DoS | 0.91 | 0.96 | 0.93 | 0.99 | 1.00 | 1.00 |
|  | Normal | 0.00 | 0.00 | 0.00 | 0.96 | 0.71 | 0.82 |
|  | Reconn | 0.99 | 1.00 | 1.00 | 1.00 | 1.00 | 1.00 |
| BLSTM | DDoS | 0.99 | 0.87 | 0.92 | 0.92 | 1.00 | 0.99 |
|  | DoS | 0.88 | 0.99 | 0.93 | 1.00 | 0.99 | 0.99 |
|  | Normal | 0.00 | 0.00 | 0.00 | 0.94 | 0.81 | 0.87 |
|  | Reconn | 0.99 | 1.00 | 1.00 | 1.00 | 0.99 | 1.00 |

Table IV shows the performance of the three models on the Bot-IoT dataset when the dataset is encoded using a classical Auto-encoder model and when the proposed preprocessing approach is deployed. For purposes of visualisation, the performance in Table IV, is broken down into the performance metrics of Precision, Recall and F1-Score and visualized in Figure 3, Figure 4, and Figure 5, respectively. From the Figures, it can be seen that, because of the insufficient training samples, the classical AE-based approaches totally fail to recognise the "normal" class instances. On the contrary, however, the proposed Transformer-based approach shows improved recognition for all attack classes with the BLSTM model recognizing the "normal" instances with values above 80% in terms of all three performance metrics.

This outstanding performance of the ViT-based approach can be attributed the fact that it deploys self-attention mechanism to learn intricate patterns by focusing on particular components of the data points for each class consequently improving its ability to discriminate between the various classes.

## V. CONCLUSION

This work demonstrated a simple yet effective approach that adapts the transformer model to work with data in form .pcap/.csv files instead of text and image data. In particular this study proposed a preprocessing technique to transform the .pcap or structured .csv traffic datasets into a format that can be processed by the ViT for classification.

In addition to adapting the ViT model to net flow packets, it was also enhanced to allow use of any classification algorithm instead of the MLP model. This was realised by stacking three different deep learning architectures including LSTM, Bidirectional LSTM and GRU models. Evaluated on two IoT traffic datasets, the proposed approach recorded competitive performance in terms of precision, recall, and F1-score for all classes from both datasets.